\pdfoutput=1

\documentclass[11pt]{article}
\usepackage{hyperref}
\usepackage{url}
\usepackage{graphicx}
\usepackage{float}
\usepackage{colortbl}
\usepackage{booktabs}
\usepackage{amsmath}
\usepackage{amssymb}
\usepackage{verbatim}
\usepackage{booktabs}
\usepackage{multirow}
\usepackage{makecell}
\usepackage{multicol}
\usepackage{tcolorbox}
\usepackage{wrapfig}
\usepackage{color}
\usepackage{url}
\usepackage{caption}
\usepackage{CJK}
\definecolor{yan}{rgb}{0.5,0.3,0.5}

\definecolor{ben}{rgb}{0.9,0.,0.5}

\definecolor{navyblue}{RGB}{191, 209, 229} 
\definecolor{light_yellow}{RGB}{255,243,194}
\definecolor{orange}{RGB}{255,200,100}
\definecolor{red}{RGB}{255, 0, 0}
\definecolor{green}{RGB}{0, 176, 80}
\newcommand{\best}{\cellcolor{orange}}

\usepackage[final]{acl}
\usepackage{multirow}
\usepackage{booktabs}
\usepackage{times}
\usepackage{latexsym}
\usepackage{amsmath}
\usepackage{amssymb}
\usepackage{algorithm}
\usepackage[T1]{fontenc}

\usepackage[utf8]{inputenc}
\usepackage{algorithmic}
\usepackage{natbib}

\usepackage{microtype}

\usepackage{inconsolata}

\usepackage{graphicx}

\title{OracleSage: Towards Unified Visual-Linguistic Understanding of Oracle Bone Scripts through Cross-Modal Knowledge Fusion}

\author{Hanqi Jiang\textsuperscript{\textmd{1,*}}, Yi Pan\textsuperscript{\textmd{1,*}}, Junhao Chen\textsuperscript{\textmd{1,*}}, Zhengliang Liu\textsuperscript{\textmd{1,*}}, Yifan Zhou\textsuperscript{\textmd{1}}, {\bf Peng Shu\textsuperscript{\textmd{1}}}, {\bf Yiwei Li\textsuperscript{\textmd{1}}}, \\
{\bf Huaqin Zhao\textsuperscript{\textmd{1}}, Stephen Mihm\textsuperscript{\textmd{1}}, Lewis C Howe\textsuperscript{\textmd{1}}, Tianming Liu\textsuperscript{\textmd{1,$\dagger$}}} \\
\textsuperscript{1}University of Georgia \\
\texttt{\textsuperscript{*}Equal Contribution}\\
\texttt{\textsuperscript{$\dagger$}Corresponding author}}

\begin{document}
\begin{CJK}{UTF8}{gbsn}
\maketitle

\begin{abstract}
Oracle bone script (OBS), as China's earliest mature writing system, present significant challenges in automatic recognition due to their complex pictographic structures and divergence from modern Chinese characters. We introduce \textbf{OracleSage}, a novel cross-modal framework that integrates hierarchical visual understanding with graph-based semantic reasoning. Specifically, we propose (1) a Hierarchical Visual-Semantic Understanding module that enables multi-granularity feature extraction through progressive fine-tuning of LLaVA's visual backbone, (2) a Graph-based Semantic Reasoning Framework that captures relationships between visual components and semantic concepts through dynamic message passing, and (3) OracleSem, a semantically enriched OBS dataset with comprehensive pictographic and semantic annotations. Experimental results demonstrate that OracleSage significantly outperforms state-of-the-art vision-language models. This research establishes a new paradigm for ancient text interpretation while providing valuable technical support for archaeological studies.
\end{abstract}

\section{Introduction}
Oracle Bone Script (OBS), as the earliest known Chinese writing system dating to the Shang Dynasty (circa 1250-1050 BCE), represents a crucial bridge to understanding ancient Chinese civilization. Traditional scholarly research has revealed its profound significance in areas ranging from philosophical concepts like truth~\cite{youngsam2015etymological} to historical figures and cultural practices. Although early studies focused primarily on cultural and philosophical interpretation, recent years have witnessed a transformation in how we approach these ancient texts, with computational methods complementing traditional scholarly expertise.

The interpretation of OBS presents a unique confluence of challenges and opportunities. The corpus encompasses over 150,000 discovered fragments containing approximately 4,500 distinct characters, where only about 1,800 have been definitively interpreted. Also, the complex variation in character forms and the limited number of expert scholars in the field have constrained the pace of interpretation using conventional methods. Recent computational methods have made significant strides, from dual-view recognition systems~\cite{9175542} to sophisticated pattern analysis~\cite{9757826}. The emergence of AI techniques, including diffusion models for character generation~\cite{guan-etal-2024-deciphering,li2023diff} and unsupervised structure-texture separation networks~\cite{9757826}, has opened new possibilities for supporting scholarly interpretation.

However, current approaches often operate in isolation, either focusing on visual pattern recognition or linguistic analysis, but rarely combining both effectively. While recent work has addressed specific challenges such as low-resource feature extraction~\cite{data6120134} and character recognition benchmarking~\cite{wang2024dataset}, there remains a critical need for approaches that can unify visual and linguistic understanding. This fragmentation limits our ability to leverage the full potential of modern computational methods in supporting comprehensive interpretation of OBS.

To address the challenging task of multi-modal OBS interpretation, we propose \textbf{OracleSage}, a multi-modal large language model framework. Departing from traditional character recognition approaches, we pioneer a dual-perspective methodology that comprehends oracle bone inscriptions through both glyph morphology and semantics, closely mirroring expert interpretation processes.

Our technical innovations are inspired by advances in hierarchical visual representation learning~\cite{he2022masked} and the successful application of heterogeneous graph neural networks in knowledge reasoning~\cite{wang2019heterogeneous}, with specific adaptations for OBS characteristics. Specifically, building upon the LLaVA~\cite{liu2023visual} vision-language architecture, we design a Hierarchical Visual-Semantic Understanding module. Through progressive fine-tuning of the visual backbone network, this module achieves multi-granularity feature extraction from overall character morphology to specific components, effectively addressing the complexity of oracle bone glyphs. Furthermore, we introduce a Graph-based Semantic Reasoning Framework that innovatively applies heterogeneous graph networks to ancient text interpretation, explicitly modeling the intricate relationships between glyph components, structural arrangements, and semantic concepts.

The main contributions of this work can be summarized as follows:

\begin{itemize}
   \item We introduce \textbf{OracleSage}, which marks the first application of multi-modal large language models in OBS interpretation and introduces hierarchical visual-semantic understanding and graph-based reasoning mechanisms for comprehensive character understanding.
   
   \item We develop \textbf{OracleSem}, the first OBS dataset with structural analysis and semantic evolution annotations, establishing a foundation for future research.
   
   \item We propose a novel paradigm for ancient text interpretation by formalizing expert interpretation processes into a computational model, providing insights for other ancient writing system studies.
\end{itemize}

This research not only advances the application of artificial intelligence in archaeology but also provides a new paradigm for domain-specific adaptation of multi-modal large language models, offering both academic value and practical significance.

\section{Related Work} 
The origins of Chinese writing can be traced back to the oracle-bone inscriptions of the Late Shang dynasty, which were written in a script ancestral to all subsequent forms of Chinese writing~\cite{10.2307/2928708,Dematte2022}. Identifying and deciphering OBS is crucial in the study of ancient China and requires a deep understanding of the culture~\cite{8978032}. Various studies have focused on improving OBS recognition accuracy through the creation of datasets with deep learning perspectives such as OBI-100~\cite{ijgi11010045}. Wang et. al.~\cite{wang2024dataset} introduced the Oracle-MNIST dataset for benchmarking machine learning algorithms in recognizing ancient characters. Additionally, the creation of datasets like HUST-OBC has been crucial in advancing research in OBS recognition and decipherment~\cite{wang2024open}. Guan et. al.~\cite{guan2024open} also introduced the EVOBC dataset, spanning six historical stages to aid in the study of the evolution of oracle bone characters. Furthermore, Li et. al.~\cite{li2024oracle} proposed the Oracle Bone Inscriptions Multi-modal Dataset (OBIMD), which provides a pixel-aligned representation of rubbings and facsimiles, alongside detailed annotations for tasks such as character recognition, rubbing denoising, and reading sequence prediction. These datasets collectively represent the progression from simpler annotation schemes to multi-modal and highly annotated datasets. However, current datasets still are deficient in semantically rich annotations that integrate multi-modal data with finer linguistic granularity. Enhancing datasets with semantically enriched and language-granular annotations would lay a crucial foundation for more efficient and accurate AI-powered OBS recognition and decipherment in the future, providing valuable insights into ancient Chinese writing systems.

Building on these datasets, recent research has explored diverse computational approaches to decipher OBS. Chang et al.~\cite{10.1145/3503161.3547925} proposed Sundial-GAN, a cascade generative adversarial networks framework that simulates the evolutionary process of Chinese characters across different historical stages. This approach addresses the challenge of directly mapping OBS to modern characters by breaking down the transformation into multiple stages, allowing for more nuanced character evolution modeling. Taking a different structural perspective, Wang et. al.~\cite{wang2024puzzle} introduced the Puzzle Pieces Picker (P3) method, which approaches decipherment through radical reconstruction. By deconstructing OBS into foundational strokes and radicals, then employing a Transformer model to reconstruct them into modern counterparts, P3 effectively captures the intricate structural relationships between ancient and modern characters. Most recently, Guan et al.~\cite{guan-etal-2024-deciphering} developed Oracle Bone Scipt Decipher (OBSD), introducing a novel approach using diffusion models for OBS decipherment. This method leverages conditional diffusion-based strategies to generate modern character forms, offering a new perspective that circumvents the limitations of traditional NLP methods in handling ancient scripts. However, these approaches primarily focus on structural and visual aspects of the characters while lacking consideration of semantic information crucial for accurate decipherment. This limitation suggests an opportunity to leverage large vison-language models~\cite{liu2024LLaVAnext,liu2023improvedLLaVA,liu2023visual} to align structural features with rich semantic information, potentially enabling more comprehensive and accurate decipherment by simultaneously considering both the visual evolution of characters and their underlying meanings in historical contexts.


\section{OracleSem}
In this work, we introduce a novel semantic-structural approach to OBS interpretation, fundamentally re-imagining how ancient Chinese characters can be understood through computational methods. Traditional approaches to OBS recognition typically rely on direct character-to-character mappings between ancient and modern Chinese scripts, which often oversimplifies the rich linguistic and cultural evolution embedded within these characters. We argue that the pictographic nature of OBS necessitates a more nuanced understanding that encompasses both structural features and semantic evolution. Our methodology is particularly significant because OBS characters, as the earliest known Chinese writing system, often exhibit complex pictographic elements that carry profound cultural and linguistic information beyond mere shape correspondence to modern characters.

\begin{figure}[t]
  \includegraphics[width=\columnwidth]{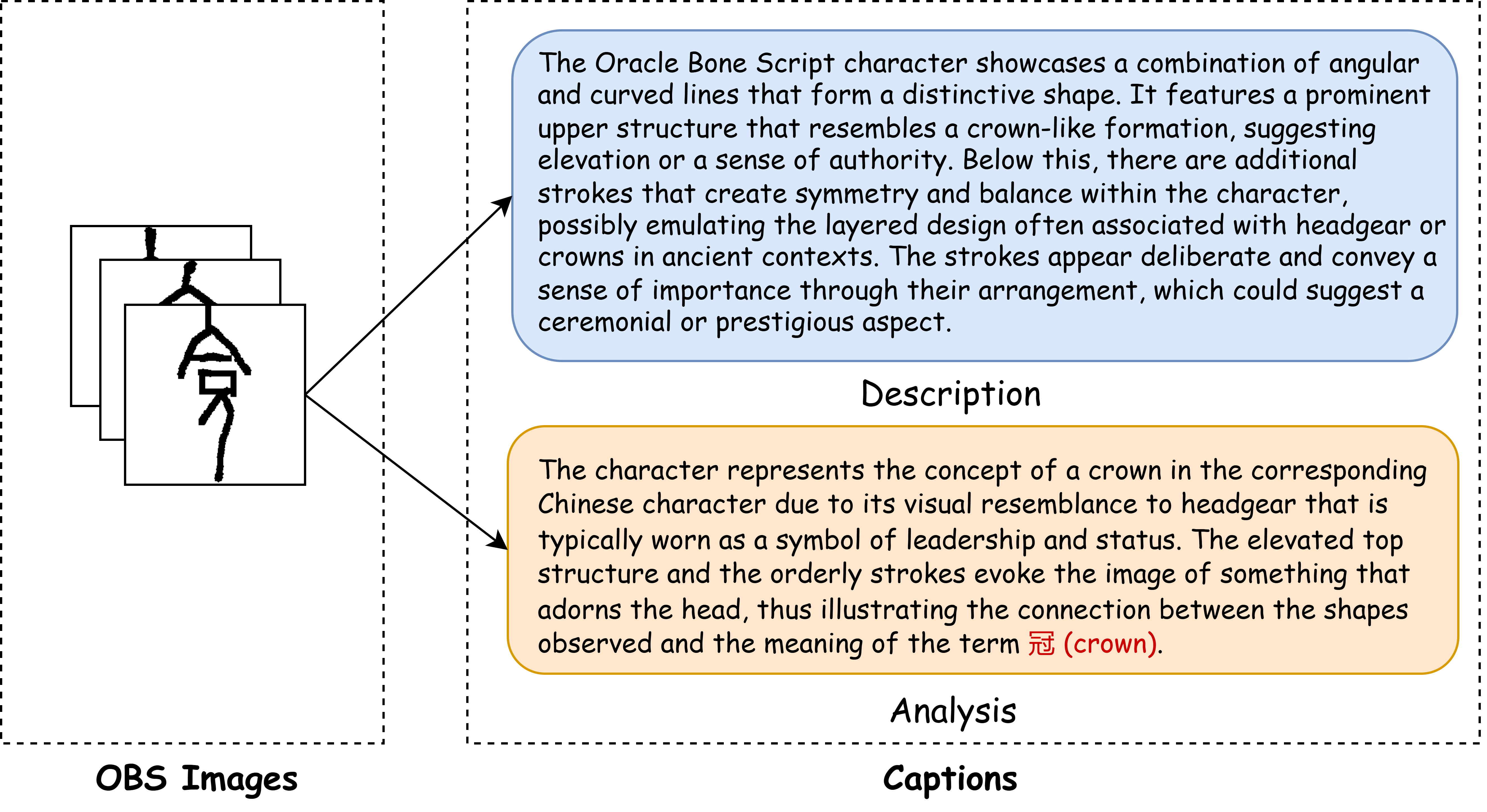}
  \caption{An example of our proposed dataset OracleSem.}
  \label{fig:dataset}
\end{figure}

To facilitate this comprehensive understanding, we present OracleSem, a semantically enriched OBS dataset based on HUST-OBS \cite{Wang2024} and EVOBC \cite{guan2024open}. OracleSem extends beyond traditional character recognition datasets by incorporating rich semantic annotations that capture multiple layers of linguistic information. For each character, OracleSem provides detailed annotations documenting its pictographic composition, structural organization, and semantic evolution pathway. These annotations systematically record the character's original pictographic meaning, its structural decomposition, and the semantic relationships that connect it to modern Chinese characters. To ensure the quality and comprehensiveness of these annotations, we employed GPT-4o to generate initial semantic descriptions, followed by rigorous verification and refinement by experts in paleography and ancient Chinese linguistics. The expanded OracleSem dataset comprises 1,762 characters, with each character associated with 10 to 20 images.

OracleSem represents a significant advancement in OBS research resources, as it bridges the gap between pure visual recognition and semantic understanding. Each character entry in OracleSem is accompanied by comprehensive documentation of its evolutionary trajectory, including intermediate forms, semantic shifts, and usage contexts. This rich semantic layer enables researchers and models to understand OBS characters not merely as visual patterns to be matched, but as complex linguistic symbols embedded within a broader historical and cultural context. Furthermore, our annotation scheme captures the polysemy that many OBS characters may have, acknowledging the scholarly debates and uncertainties that often surround ancient text interpretation.

\section{Method}

\subsection{Preliminary}
Our work builds upon LLaVA-1.5~\cite{liu2023improvedLLaVA} a state-of-the-art vision-language model that seamlessly integrates visual understanding with language generation capabilities. The LLaVA architecture consists of three fundamental components: a CLIP-based vision encoder for visual feature extraction, a Q-Former for multi-modal alignment, and a Llama-based language decoder for text generation.

The visual perception pathway begins with the CLIP vision encoder, which transforms input images into rich feature representations through a series of transformer blocks. Specifically, given an input image $\mathbf{x}$, the vision encoder $E_v$ maps it to a sequence of visual features $\mathbf{V} \in \mathbb{R}^{P \times d}$, where $P$ denotes the number of image patches and $d$ represents the feature dimension. These features form a hierarchical representation that captures both fine-grained visual details and high-level semantic patterns, essential for understanding the intricate structures of oracle bone characters.

The Q-Former module serves as a crucial bridge between visual and linguistic modalities \cite{wang2022image}. Through a set of learnable queries $\mathbf{Q} \in \mathbb{R}^{M \times d}$, where $M$ is the number of queries, it adaptively extracts task-relevant visual information and projects it into a shared semantic space. This cross-modal projection is achieved through multi-head attention mechanisms:
\begin{equation}
\mathbf{H} = \text{MultiHead}(\mathbf{Q}, \mathbf{K}, \mathbf{V})
\end{equation}
\begin{equation}
\mathbf{Z} = \text{FFN}(\mathbf{H})
\end{equation}
where $\mathbf{Z}$ represents the aligned multi-modal features that bridge the visual-linguistic gap.

For language understanding and generation, LLaVA employs a modified Llama architecture \cite{touvron2023llama}, which has demonstrated remarkable capabilities in processing and generating natural language. This decoder component takes the aligned multi-modal representations and generates contextually appropriate textual outputs through an autoregressive process. The architectural design of Llama, with its enhanced attention mechanisms and scaled-up model capacity, makes it particularly suitable for our task of OBS interpretation, where both linguistic expertise and cultural context understanding are essential.

\subsection{Hierarchical Visual-Semantic Understanding}
\begin{figure*}[t]
  \includegraphics[width=\linewidth]{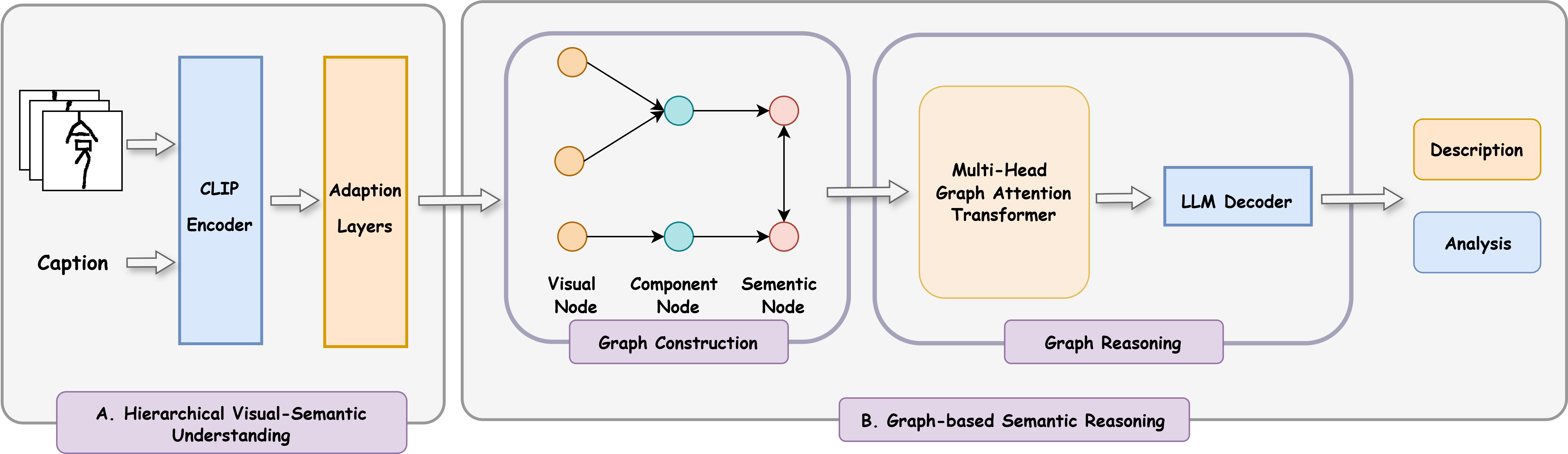} 
  \caption {Our purposed fine-tuning framework.}
\end{figure*}

To address the challenge of recognizing OBS with limited data, we propose a Hierarchical Visual-Semantic Understanding (HVSU) module. This module extends the visual backbone of LLaVA by incorporating specialized structures tailored to the unique characteristics of OBS while preserving pre-trained knowledge through careful fine-tuning.

The HVSU module introduces lightweight adaptation layers atop the frozen CLIP visual encoder. Given an input feature matrix $\mathbf{X}$, the adaptation layers are formulated as:
\begin{align}
\mathbf{X}' &= \text{LN}\left(\mathbf{X} + \text{Dropout}\left(\text{MHA}(\mathbf{X})\right)\right) \\
\mathbf{X}_{\text{out}} &= \text{LN}\left(\mathbf{X}' + \text{Dropout}\left(\text{FFN}(\mathbf{X}')\right)\right)
\end{align}
where $\text{LN}$ denotes layer normalization, $\text{MHA}$ represents multi-head self-attention, and $\text{FFN}$ is a feed-forward network.

Operating on a three-level hierarchy, the HVSU module first enhances character-level representations. The standard CLIP patch embedding is augmented with adaptive pooling to produce multi-scale feature maps $\{\mathbf{F}_i\}_{i=1}^n$, where each $\mathbf{F}_i \in \mathbb{R}^{H_i \times W_i \times d}$ captures features at different scales. This design enables the model to grasp both global structures and fine details critical for OBS recognition.

At the component level, a lightweight region proposal mechanism identifies potential semantic components within characters. For each region $\mathbf{r}_i$, attention-weighted features are computed as:

\begin{equation}
\mathbf{Q}_i = \mathbf{F}_{\mathbf{r}_i} \mathbf{W}_q, \quad
\mathbf{K}_i = \mathbf{F}_{\mathbf{r}_i} \mathbf{W}_k, \quad
\mathbf{V}_i = \mathbf{F}_{\mathbf{r}_i} \mathbf{W}_v
\end{equation}

\begin{equation}
\mathbf{A}_i = \text{Softmax}\left(\frac{\mathbf{Q}_i \mathbf{K}_i^\top}{\sqrt{d}}\right) \mathbf{V}_i
\end{equation}

\noindent where $\mathbf{F}_{\mathbf{r}_i}$ are the features corresponding to region $\mathbf{r}_i$, and $\mathbf{W}_q, \mathbf{W}_k, \mathbf{W}_v$ are learnable projection matrices. This mechanism focuses on salient components contributing to the semantic meaning of the character.

To model structural relationships among components, a Spatial Relation Encoder (SRE) represents OBS as directed graphs $\mathcal{G} = (\mathcal{V}, \mathcal{E})$, where nodes $\mathcal{V}$ correspond to components and edges $\mathcal{E}$ encode spatial relationships. Structural features are processed through graph attention layers:
\begin{equation}
\mathbf{H}' = \text{GAT}(\mathbf{H}, \mathbf{A})
\end{equation}
where $\mathbf{H}$ is the node feature matrix, $\mathbf{A}$ is the adjacency matrix derived from spatial relationships, and $\text{GAT}$ denotes the graph attention network.

Finally, a Dynamic Feature Fusion Module (DFFM) adaptively integrates hierarchical features. For each component, a gating mechanism determines the contribution of different feature levels:
\begin{align}
\mathbf{g}_i &= \sigma\left(\mathbf{W}_g \left[\mathbf{F}_{\mathbf{r}_i} \parallel \mathbf{A}_i \parallel \mathbf{H}_i\right]\right) \\
\mathbf{F}_{\text{fused}} &= \sum_i \mathbf{g}_i \odot \text{Conv}\left(\mathbf{F}_{\mathbf{r}_i} \parallel \mathbf{A}_i \parallel \mathbf{H}_i\right)
\end{align}
where $\sigma$ is the sigmoid activation, $\mathbf{W}_g$ is a learnable weight matrix, $\parallel$ denotes concatenation, and $\odot$ represents element-wise multiplication. This fusion mechanism dynamically adjusts the contributions of different feature levels based on their relevance, enhancing the model's ability to interpret complex OBS.

\subsection{Graph-based Semantic Reasoning Framework}

Building upon the features extracted by the HVSU module, we introduce a Graph-based Semantic Reasoning Framework (GSRF) to exploit the inherent graph-like nature of OBS. The framework constructs a heterogeneous graph $\mathcal{G} = (\mathcal{V}, \mathcal{E})$, where nodes $\mathcal{V}$ include visual nodes $\mathbf{v}_i$, component nodes $\mathbf{c}_j$, and semantic nodes $\mathbf{s}_k$, and edges $\mathcal{E}$ encompass spatial relationships ($\mathcal{E}_{cc}$), visual-component associations ($\mathcal{E}_{vc}$), and component-semantic connections ($\mathcal{E}_{cs}$).

A hierarchical message passing mechanism updates node features through:

\begin{equation}
\mathbf{m}_{i,j} = \text{MLP}\left(\left[\mathbf{h}_i^t \parallel \mathbf{h}_j^t \parallel \mathbf{e}_{i,j}\right]\right)
\end{equation}

\begin{equation}
\resizebox{0.85\linewidth}{!}{$
\alpha_{ij} = \frac{\exp\left(\text{LeakyReLU}\left(\mathbf{a}^\top \left[\mathbf{W} \mathbf{h}_i^t \parallel \mathbf{W} \mathbf{h}_j^t\right]\right)\right)}{\sum_{k \in \mathcal{N}(i)} \exp\left(\text{LeakyReLU}\left(\mathbf{a}^\top \left[\mathbf{W} \mathbf{h}_i^t \parallel \mathbf{W} \mathbf{h}_k^t\right]\right)\right)}
$}
\end{equation}

\begin{equation}
\mathbf{h}_i^{t+1} = \text{GRU}\left(\mathbf{h}_i^t, \sum_{j \in \mathcal{N}(i)} \alpha_{ij} \mathbf{m}_{i,j}\right)    
\end{equation}

where $\mathbf{h}_i^t$ is the feature vector of node $i$ at time $t$, $\mathbf{e}_{i,j}$ represents edge features, $\text{MLP}$ is a multi-layer perceptron, $\mathbf{a}$ and $\mathbf{W}$ are learnable parameters, and $\mathcal{N}(i)$ denotes the neighboring nodes of $i$. The attention coefficients $\alpha_{ij}$ capture the relative importance of neighboring nodes, facilitating effective information aggregation.

To adapt to evolving relationships during reasoning, we introduce a dynamic graph update mechanism:
\begin{equation}
\mathbf{E}^{t+1} = \text{Update}\left(\mathbf{E}^t, \mathbf{H}^t, \mathcal{L}^t\right),
\end{equation}
where $\mathbf{E}^{t+1}$ represents the updated edge set based on the current node features $\mathbf{H}^t$ and accumulated loss $\mathcal{L}^t$. This mechanism allows the framework to dynamically adjust the graph structure, aligning visual, structural, and semantic representations to enhance the interpretation of OBS.

\begin{algorithm}
\caption{OracleSage Model}
\begin{algorithmic}
\REQUIRE Oracle bone image $x$, pre-trained LLaVA model $\theta_{LLaVA}$
\ENSURE Semantic interpretation $y$

\STATE // Hierarchical Visual Feature Extraction
\STATE $F \leftarrow \text{CLIPEncoder}(x)$
\STATE $F_{pyramid} \leftarrow \text{GenerateFeaturePyramid}(F)$
\STATE $R \leftarrow \text{RegionProposal}(F_{pyramid})$
\STATE $S \leftarrow \text{SpatialEncoder}(R)$
\STATE $H \leftarrow \text{FeatureFusion}(F_{pyramid}, R, S)$

\STATE // Graph-based Semantic Reasoning
\STATE Initialize graph $G$ with nodes $V = \{v_i, c_j, s_k\}$ and edges $E = \{E_{cc}, E_{vc}, E_{cs}\}$
\FOR{$t = 1$ to $T$}
    \FOR{each edge $(i,j) \in E$}
        \STATE $m_{ij} \leftarrow \text{MLP}([h_i; h_j; e_{ij}])$
        \STATE $\alpha_{ij} \leftarrow \text{Attention}(h_i, h_j)$
        \STATE $h_i \leftarrow \text{GRU}(h_i, \sum_{j \in \mathcal{N}(i)} \alpha_{ij} m_{ij})$
    \ENDFOR
    \STATE $G \leftarrow \text{UpdateGraph}(G)$
\ENDFOR

\STATE $y \leftarrow \text{LLMDecoder}(G)$
\RETURN $y$
\end{algorithmic}
\end{algorithm}

\subsection{Training Procedure and Loss Function}

We employ a progressive fine-tuning strategy to train the proposed model, integrating the HVSU and GSRF modules. Initially, the CLIP visual encoder is kept frozen while training the adaptation layers in the HVSU module to capture script-specific features without distorting pre-trained representations. Next, we fine-tune the last two transformer blocks of the CLIP encoder along with the adaptation layers to better adapt the model to the domain of OBS. Finally, we jointly train the entire network, including the HVSU and GSRF modules, using a small learning rate of $5 \times 10^{-6}$ to achieve optimal performance.

The total loss function is designed to balance multiple objectives, ensuring comprehensive learning. It is defined as:
\begin{equation}
\mathcal{L}_{\text{total}} = \lambda_1 \mathcal{L}_{\text{char}} + \lambda_2 \mathcal{L}_{\text{comp}} + \lambda_3 \mathcal{L}_{\text{struct}} + \lambda_4 \mathcal{L}_{\text{sem}},
\end{equation}
where $\lambda_1, \lambda_2, \lambda_3$, and $\lambda_4$ are weighting hyperparameters determined empirically. The character-level loss $\mathcal{L}_{\text{char}}$ ensures accurate recognition of oracle bone characters:
\begin{equation}
\mathcal{L}_{\text{char}} = \text{CrossEntropy}(\mathbf{y}_{\text{char}}, \hat{\mathbf{y}}_{\text{char}}),
\end{equation}
where $\mathbf{y}_{\text{char}}$ and $\hat{\mathbf{y}}_{\text{char}}$ are the ground truth and predicted character labels. The component-level loss $\mathcal{L}_{\text{comp}}$ focuses on accurate identification of semantic components:
\begin{equation}
\mathcal{L}_{\text{comp}} = \sum_{i} \text{CrossEntropy}(\mathbf{y}_{\text{comp},i}, \hat{\mathbf{y}}_{\text{comp},i}),
\end{equation}
with $\mathbf{y}_{\text{comp},i}$ and $\hat{\mathbf{y}}_{\text{comp},i}$ being the ground truth and predicted labels for component $i$. The structural consistency loss $\mathcal{L}_{\text{struct}}$ captures spatial relationships among components:
\begin{equation}
\mathcal{L}_{\text{struct}} = \|\mathbf{H}^{t+1} - \text{GNN}(\mathbf{H}^t, \mathbf{E}^t)\|_2^2 + \beta \|\mathbf{A} - \hat{\mathbf{A}}\|_1,
\end{equation}
where $\mathbf{H}^{t+1}$ is the updated node feature matrix, $\text{GNN}$ denotes the graph neural network operation in the GSRF module, $\mathbf{A}$ is the ground truth adjacency matrix, $\hat{\mathbf{A}}$ is the predicted adjacency matrix, and $\beta$ is a weighting factor. Finally, the semantic alignment loss $\mathcal{L}_{\text{sem}}$ encourages effective mapping between visual features and semantic concepts:
\begin{equation}
\mathcal{L}_{\text{sem}} = \sum_{k} \text{CrossEntropy}(\mathbf{y}_{\text{sem},k}, \hat{\mathbf{y}}_{\text{sem},k}),
\end{equation}
where $\mathbf{y}_{\text{sem},k}$ and $\hat{\mathbf{y}}_{\text{sem},k}$ are the ground truth and predicted semantic labels for concept $k$. By integrating these losses, the training process ensures that the model not only recognizes oracle bone characters accurately but also captures component structures and semantic alignments, facilitating a comprehensive understanding of the scripts.




\section{Experiments}

\subsection{Dataset and Evaluation Setup}
We evaluate OracleSage on our proposed OracleSem dataset. Our OracleSem dataset enriches these characters with comprehensive semantic annotations, including pictographic interpretations, component analysis, and modern Chinese character mappings. We maintain the standard 9:1 train-test split ratio, ensuring that test set characters are completely disjoint from the training set to evaluate the model's generalization capability.
\subsection{Implementation Details}
We implement OracleSage based on LLaVA-13B architecture. For the HVSU module, we initialize the visual encoder with CLIP ViT-L/14 weights and add three adaptation layers. The graph reasoning module is implemented with 3 reasoning steps and 8 attention heads. During training, we employ a three-phase progressive fine-tuning strategy with a learning rate of 5e-6 and AdamW optimizer. All experiments are conducted on 8 NVIDIA A5000 GPUs with 24GB memory each. 
\subsection{Quantitative Results}

As shown in Table\ref{tab:comparison}, the evaluation of Oracle character recognition methods reveals a clear distinction in performance across Diffusion-based, Deep Learning-based, and LLM-based approaches. Among Diffusion-based methods, OBSD achieved moderate success, while CycleGAN \cite{zhu2017cyclegan} failed entirely in both metrics. Deep Learning-based methods, particularly ViT \cite{vaswani2017attention} and ResNet-50 \cite{he2016resnet}, delivered the highest accuracy, with ViT slightly outperforming ResNet-50. In contrast, LLM-based methods faced challenges in achieving high accuracy but provided a foundation for integrating reasoning and explanation into predictions. 

For other LLM-based methods, such as GPT-4o \cite{achiam2023gpt} and Gemini \cite{team2023gemini}, we conducted experiments by providing several example descriptions and analyses using ground truth data for randomly selected OBS images. These examples served as a few-shot learning prompt to evaluate the models' ability to analyze and recognize Oracle characters. Despite this additional context and guidance, these models struggled to generalize effectively to the recognition task. Their inability to extract meaningful patterns from Oracle characters shows the limitations of these models in handling the unique challenges posed by the intricate and low-resource nature of Oracle script data. Notably, our proposed method, OracleSage, demonstrated significant improvement over other LLM-based methods in both Top-1 and Top-10 accuracy.

\begin{table}[t]
   \caption{Performance comparison of different methods. Accuracy (ACC [\%]) is reported for evaluation.}
   \centering
   \resizebox{0.95\linewidth}{!}{
   \begin{tabular}{lcc}
       \toprule
       \multirow{2}{*}{Method} & \multicolumn{2}{c}{Accuracy [\%]} \\
       \cmidrule{2-3}
       & Top-1@Acc & Top-10@Acc \\
       \midrule
       \textit{Diffusion-based Method} \\
       CycleGAN \cite{zhu2017cyclegan} & 0.0\% & 0.0\% \\
       BBDM \cite{li2023bbdm} & 17.3\% & 29.6\% \\
       OBSD \cite{guan-etal-2024-deciphering} & 39.4\% & 48.5\% \\
       \hline \hline
       \textit{Deep Learning-based Method} \\
       AlexNet \cite{krizhevsky2012alexnet} & 5.11\% & 23.7\% \\
       ResNet-50 \cite{he2016resnet} & 88.7\% & 90.3\% \\
       ViT \cite{vaswani2017attention} & 90.1\% & 92.4\% \\
       \hline \hline
       \textit{LLM-based Method} \\
       GPT-4o~\cite{achiam2023gpt} & 3.0\% & 3.0\% \\
       Gemini~\cite{team2023gemini} & 0.0\% & 0.0\% \\
       Pixtral-12B \cite{agrawal2024pixtral} & 0.0\% & 0.0\% \\
       \best {OracleSage-13B (Ours)} & \best 20.2\% & \best 40.9\% \\
       \bottomrule
   \end{tabular}
   }
   \label{tab:comparison}
   \vspace{-0.65cm}
\end{table}

\begin{figure*}[t!]
  \centering
  \resizebox{0.8\linewidth}{!}{%
    \includegraphics{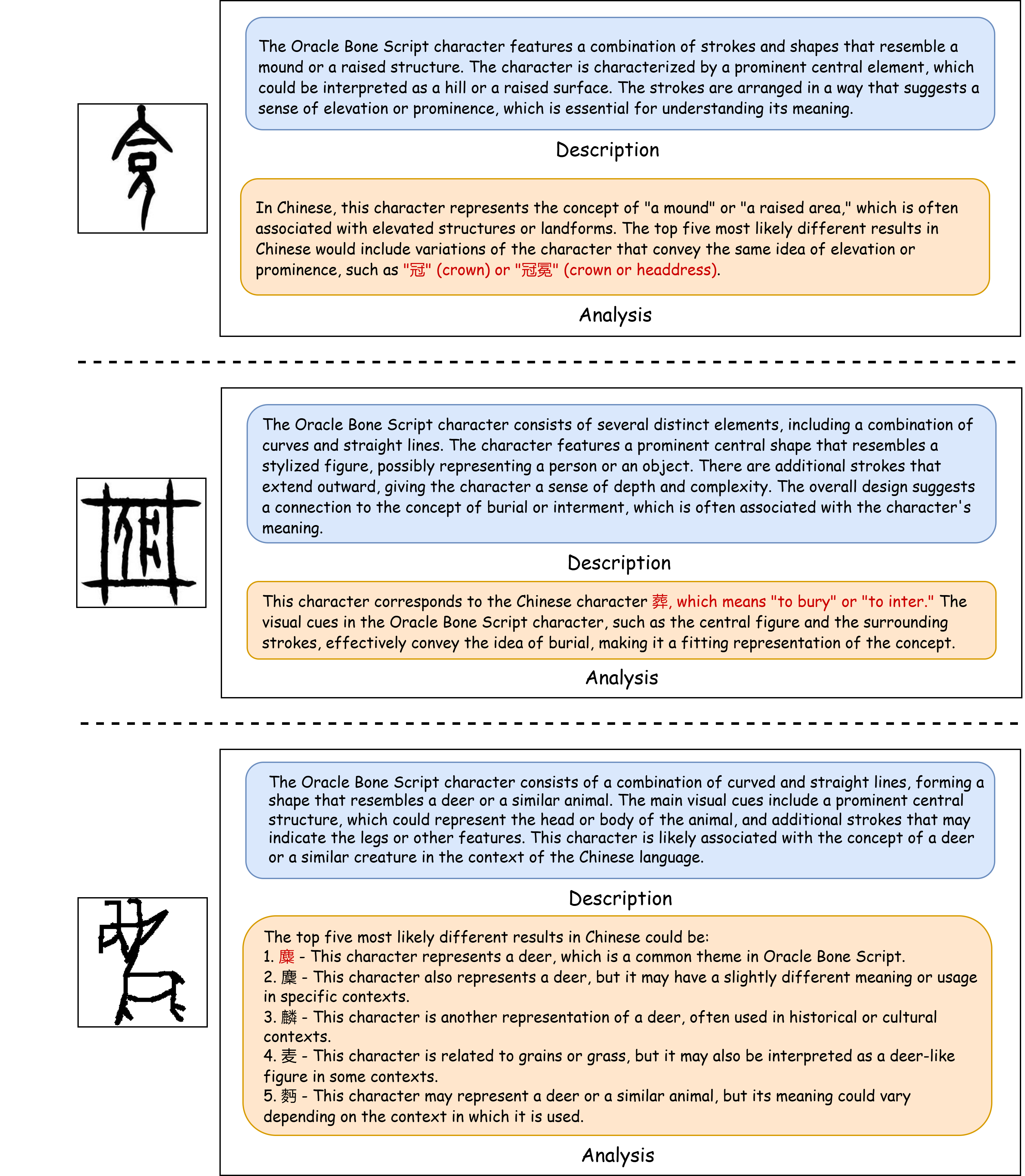}
  }
  \caption{Examples demonstrating our method's ability to infer and recognize oracle images across three test cases.}
  \label{fig:oracle_example}
\end{figure*}

Although this evaluation shows the relative limitations of our proposed model, OracleSage, in accuracy, compared to state-of-the-art Deep Learning and Diffusion-based methods, which achieve higher accuracy, the Deep Learning and Diffusion-based methods are inherently limited in their applicability to Oracle character research. Deep Learning and Diffusion-based methods, providing high accuracy but no meaningful interpretability, which is critical for Oracle character recognition. The biggest challenge in this field is that a significant number of Oracle characters remain unrecognized, and simply predicting these characters without context or explanation offers little value to researchers. Additionally, Deep Learning and Diffusion-based methods are prone to potential overfitting, particularly when trained on limited datasets like Oracle characters, which can result in models that generalize poorly to unseen data. In contrast, OracleSage adopts a novel semantic-structural approach that combines hierarchical visual-semantic understanding with graph-based semantic reasoning. This methodology enables OracleSage to predict Oracle characters while also providing an interpretable explanation of their pictographic and semantic evolution. This focus on structural and semantic relationships not only facilitates better understanding for researchers but also addresses the broader challenges of deciphering unrecognized Oracle characters. Despite lagging behind Deep Learning-based methods in raw accuracy, OracleSage bridges the gap between visual recognition and semantic interpretation, making it a valuable tool for advancing Oracle script studies.

\subsection{Qualitative Analysis}
To provide insights into how OracleSage interprets oracle bone inscriptions, we present three representative examples in Figure \ref{fig:oracle_example}. Each case demonstrates the model's capability in different challenging scenarios. 

The first two examples demonstrate the model's ability to understand the spatial concepts inherent in Oracle Bone Script characters. In the first case, the character features an up-and-down structure resembling a crown placed on a human's head. The arrangement of elements, with the "crown" positioned above and the "head" below, conveys the concept of elevation and prominence, which is crucial to understanding its meaning. In Chinese, this character corresponds to the idea of "a crown" or "a raised area," symbolizing elevated structures or headpieces. Likely interpretations include variations such as "冠" (crown) or "冠冕" (crown or headdress), reflecting its spatial arrangement.

In the second case, the character showcases a more complex spatial arrangement of curves and straight lines, forming a central shape that resembles a stylized figure, possibly representing a person or object. Additional outward-extending strokes create a sense of depth and complexity, indicative of the concept of burial or interment. This character corresponds to "葬" in Modern Chinese, meaning "to bury" or "to inter." The spatial composition, including the central figure and surrounding strokes, effectively conveys its semantic connection.

The third example highlights a character formed by curved and straight lines resembling a deer or similar animal. OracleSage identifies the top five most likely results in Chinese, which include "麋," "麇," "麟," "麦," and "麪." Notably, "麇" and "麟" emerge as reasonable inferences, as they also represent deer or deer-like concepts in historical and cultural contexts. The ability to differentiate these plausible interpretations demonstrates the model's robust reasoning and nuanced understanding of the Oracle Bone Script.
These examples collectively showcase OracleSage's ability to grasp spatial and semantic relationships in Oracle Bone Script, highlighting its effectiveness in deciphering diverse and complex characters.

\section{Discussion}

OracleSage introduces a novel framework for interpreting Oracle Bone Script through the integration of hierarchical visual-semantic understanding with graph-based semantic reasoning. Our experiments demonstrate the effectiveness of this approach in bridging visual features with semantic understanding. The results are interpretable and align well with expert analysis processes. However, several key challenges remain to be addressed. These limitations provide important directions for future research in ancient script interpretation.

\subsection{Limitations}
OracleSage faces several challenges in its current form. First, this vision-language framework's accuracy metrics are lower than traditional deep learning approaches. This indicates a persistent gap between semantic understanding and precise character recognition.

The framework also faces important data and expert annotation constraints. Our OracleSem dataset includes only 1,762 characters. This represents a small portion of the over 150,000 discovered Oracle Bone fragments. In addition, the dataset expansion requires extensive expert involvement for annotation verification. This creates a fundamental challenge that can only be addressed through increased collaboration with linguistic experts and archaeologists.

Finally, OracleSage shows architectural constraints in its cross-script applications. The framework works well for pictographic and logographic scripts with clear visual-semantic mappings, such as Egyptian hieroglyphs or early Sumerian cuneiform. However, it requires significant modifications for other writing systems. Syllabic scripts like Linear B or abstract symbolic systems often lack direct visual-semantic correlations. This means the current architecture cannot be directly applied to these systems without fundamental changes to handle different linguistic relationships.

\subsection{Future Directions}
Despite these limitations, we identify several promising avenues for future development of OracleSage. These opportunities span both methodological improvements to the core framework and novel applications in archaeology and education. We highlight the following key directions that could advance the practical impact of this framework.

\begin{itemize}
    \item \textbf{Expanding OracleSem Dataset:} Future work can enhance OracleSem by incorporating annotations for ambiguous or partially recognized characters. Regional and temporal variations could also be added to support broader generalization across archaeological findings.

    \item \textbf{Interactive Tools for Archaeologists:} OracleSage can be extended into an interactive platform where experts refine predictions and structural breakdowns. Features such as semantic evolution visualizations can help archaeologists directly improve the model and validate its interpretations.

    \item \textbf{Educational Applications:} OracleSage's ability to align visual and linguistic reasoning can facilitate the development of tools for teaching ancient writing systems. These tools would allow students and researchers to explore structural and semantic nuances in a more hands-on manner.

    \item \textbf{Audio Integration:} Audio data, such as reconstructed ancient pronunciations, could be added to OracleSage. Expanding the framework to include vision, language, and audio would improve its ability to provide contextual insights and make accurate predictions.

    \item \textbf{Broader Applications Beyond OBS:} The current OracleSage framework is well-suited for ancient scripts with strong pictographic features and clear visual-semantic relationships. This includes Egyptian hieroglyphs, early Sumerian cuneiform, and early Chinese Bronze inscriptions. However, the framework requires fundamental modifications for writing systems without these characteristics. Syllabic scripts like Linear B and alphabetic systems would need different reasoning mechanisms to focus on phonetic relationships rather than pictographic features. We propose to develop a modular architecture where the base visual recognition remains consistent, but the semantic reasoning components can be specialized for different script types. This would allow OracleSage to maintain its strengths in pictographic analysis while extending its capabilities to a broader range of ancient writing systems.

    \item \textbf{Enhanced Interpretability:} OracleSage could improve interpretability by clearly visualizing the contributions of structural features and semantic relationships to its predictions. Providing confidence scores for uncertain outputs would offer additional clarity for researchers.

    \item \textbf{Integration with Knowledge Graphs:} Incorporating knowledge graphs would enable OracleSage to encode relationships between characters, semantic concepts, and historical contexts. This approach could facilitate richer reasoning and provide links between inscriptions and related artifacts or texts.
\end{itemize}

\section{Conclusion}
This paper introduces OracleSage, a novel framework that combines hierarchical visual-semantic understanding and graph-based semantic reasoning to address the complexities of Oracle Bone Script interpretation. By bridging visual and linguistic modalities, OracleSage not only achieves competitive recognition accuracy but also provides interpretable insights into the structural and semantic evolution of ancient characters. The development of the OracleSem dataset further enriches the research landscape with semantically annotated data, fostering advancements in computational paleography. Moreover, beyond recognizing known Oracle characters, this framework offers Oracle researchers valuable tools and inspiration to decipher discovered yet unresolved Oracle scripts, enabling deeper exploration and understanding of these ancient texts.


\bibliography{custom}



\end{CJK}
\end{document}